\documentclass{elsarticle}

\usepackage{stfloats}
\usepackage{graphicx}
\usepackage{bm}
\usepackage[mathlines]{lineno}

\usepackage{amsmath, amsfonts, amsbsy, amssymb, amsthm}
\usepackage[utf8]{inputenc}
\usepackage[T1]{fontenc}
\usepackage{mathptmx}

\usepackage{array,url,kantlipsum}
\usepackage{multicol,lipsum,xparse}

\usepackage{url}
\usepackage{setspace}
\usepackage[table]{xcolor}
\usepackage{verbatim}
\usepackage[export]{adjustbox}
\usepackage{subcaption}
\usepackage{accents} 
\usepackage{float}

\usepackage{color, colortbl}

\definecolor{c1}{RGB}{92,89,191}
\definecolor{c2}{RGB}{191,92,98}
\definecolor{c3}{RGB}{98,191,92}

\usepackage[utf8]{inputenc}
\usepackage{multirow}
\usepackage{tikz}
\usetikzlibrary{arrows,calc,positioning}
\tikzstyle{intt}=[draw,text centered,minimum size=9em,text width=5.5cm,text height=0.34cm]
\tikzstyle{intl}=[draw,text centered,minimum size=2em,text width=2.75cm,text height=0.34cm]
\tikzstyle{int}=[draw,minimum size=2.0em,text centered,text width=4.5cm]
\tikzstyle{intgi}=[draw,minimum size=4em,text centered,text width=5.2cm]
\tikzstyle{intg}=[draw,minimum size=4em,text centered,text width=5.5cm]
\tikzstyle{intgg}=[draw,minimum size=4em,text centered,text width=4.0cm]
\tikzstyle{sum}=[draw,shape=circle,inner sep=2pt,text centered,node distance=3.5cm]
\tikzstyle{summ}=[drawshape=circle,inner sep=4pt,text centered,node distance=3.cm]
\usepackage[a4paper, total={6in, 8in}]{geometry}
\usepackage{dsfont}
\usepackage{stmaryrd}
\usepackage{natbib}
\begin{document}

\title{Pretrained LLMs as Real-Time Controllers for Robot Operated Serial Production Line} 

\author{Muhammad Waseem$^1$}

\author{Kshitij Bhatta$^2$}

\author{Chen Li$^3$}

\author{Qing Chang$^{*4}$}

\address{Department of Mechanical and Aerospace Engineering, University of Virginia, Charlottesvile, VA, 22903, USA\\Email: $^1$kqr5pu@virginia.edu, $^2$qpy8hh@virginia.edu, $^3$cl4kv@virginia.edu, $^4$qc9nq@virginia.edu}
\address{$^*$ Corresponding Author}

\begin{abstract}
The manufacturing industry is undergoing a transformative shift, driven by cutting-edge technologies like 5G, AI, and cloud computing. Despite these advancements, effective system control, which is crucial for optimizing production efficiency, remains a complex challenge due to the intricate, knowledge-dependent nature of manufacturing processes and the reliance on domain-specific expertise. Conventional control methods often demand heavy customization, considerable computational resources, and lack transparency in decision-making. In this work, we investigate the feasibility of using Large Language Models (LLMs), particularly GPT-4, as a straightforward, adaptable solution for controlling manufacturing systems, specifically, mobile robot scheduling. We introduce an LLM-based control framework to assign mobile robots to different machines in robot assisted serial production lines, evaluating its performance in terms of system throughput. Our proposed framework outperforms traditional scheduling approaches such as First-Come-First-Served (FCFS), Shortest Processing Time (SPT), and Longest Processing Time (LPT). While it achieves performance that is on par with state-of-the-art methods like Multi-Agent Reinforcement Learning (MARL), it offers a distinct advantage by delivering comparable throughput without the need for extensive retraining. These results suggest that the proposed LLM-based solution is well-suited for scenarios where technical expertise, computational resources, and financial investment are limited, while decision transparency and system scalability are critical concerns.
\end{abstract}

\maketitle
\section{Introduction}
The rapid advancement of automation in industrial systems has led to increasingly complex and dynamic manufacturing environments. This complexity escalates further with even small changes to system components, such as the addition of a new robot, product type, or machine \cite{manuf_complexity}. In these cases, efficiently managing and controlling the system becomes significantly more challenging, especially when there is a need for flexibility, such as producing a new product variant or introducing a new process. As the number of system entities increases, the level of complexity grows exponentially, turning the problem into an NP-hard challenge \cite{np_hard}.

Current literature has explored various control strategies for complex manufacturing systems, including approaches based on heuristics \cite{heuristic_metaheurisitics}, meta-heuristics \cite{heuristic_metaheurisitics}, machine learning (ML) \cite{machineLearningControl}, reinforcement learning (RL) \cite{Muhammad_ASME}, and deep reinforcement learning (DRL) \cite{waseem2024nash}. It is well-established that RL and DRL-based control methods can deliver impressive efficiency and optimal performance \cite{bhatta2023integrating}. However, achieving such performance comes at a significant cost. These advanced approaches often require specialized domain expertise and substantial computational resources, which may not always be readily available or financially feasible. Furthermore, the computational burden increases when system changes occur, requiring model retraining to maintain optimal performance. This challenge is compounded by issues of scalability, where models often need significant adjustments and retraining as the system evolves. Even if optimal control is achieved, these "black-box" models typically lack transparency, making it difficult to understand the rationale behind individual decisions. While there has been some research aimed at improving explainability \cite{explainable_RL,2021explainability}, it remains a complex issue, particularly for operators who lack expertise. For these users, the decisions made by these models often remain opaque and difficult to interpret.

To address these gaps, this paper explores the potential of LLM-based control as an alternative or a complementary enhancement to existing control approaches. LLMs, which have rapidly gained attention in recent years, are evolving at a fast pace and are increasingly applied across a variety of fields, including manufacturing \cite{llmManufacturing}. The advanced natural language processing (NLP) capabilities of LLMs enhance transparency, making it easier for users to interpret and act upon decisions. Unlike RL and DRL, which require frequent retraining and hyperparameter tuning, LLMs are more adaptable to dynamic environments, offering a significant reduction in computational overhead. This makes LLMs particularly well-suited for complex manufacturing environments, where system changes and real-time decision-making are constant. 

While LLMs offer significant potential and are capable of delivering enhanced solutions, utilizing them effectively is not without its challenges. It goes beyond merely providing natural language instructions and obtaining the desired output. One of the key limitations of LLMs is their propensity for hallucination, meaning that even with clear and detailed instructions, the model may not always produce the expected or accurate results. This issue becomes even more pronounced when dealing with complex systems such as manufacturing environments where numerous parameters must be monitored, and multiple conditions need to be checked before each decision is made. In these scenarios, the risk of errors is heightened. 


To address these challenges, we propose a novel control framework for robot-operated serial production lines that combines optimal performance with interpretability. Our approach employs a structured process for action generation, leveraging LLMs to deliver human-readable explanations for each action. This enhances transparency and user understanding, addressing a key challenge in real-time control systems. The framework is scalable, adaptable to diverse production configurations, and eliminates the need for retraining or hyperparameter tuning.

To validate its feasibility and effectiveness, we evaluate the proposed framework against several baselines, including DRL-based control, FCFS, SPT, and LPT strategies. The results demonstrate that a pretrained LLM can be effectively applied to real-time control problems, achieving comparable performance in operational efficiency and decision-making accuracy, particularly in dynamic production environments.

The remainder of the paper is structured as follows: Section 2 reviews relevant literature and background, Section 3 details our methodology for evaluating LLM-based control, Section 4 present case studies and discusses our findings followed by conclusion in Section 5.

\section{Literature Review and Background}
\subsection{Literature Review}
Manufacturing systems face increasing complexity, dynamic environmental conditions, and uncertain market demands \cite{Chen_demand,manuf_complexity}. To remain competitive, manufacturers must not only deliver customized products at a low cost but also ensure rapid time-to-market. This scenario presents significant challenges for businesses striving to maximize efficiency and maintain an edge in a highly competitive landscape. To address these challenges, companies must be flexible enough to manage fluctuating demands and effectively control their production systems. Control can range from optimizing product mixes to managing machine operations or coordinating material handling systems. Effectively controlling such dynamic systems is a complex task, which has been extensively addressed in the literature from various perspectives \cite{bhatta2023integrating,bhatta2025enhancing,bhatta2024dynamic}.

Control approaches in the literature vary significantly, ranging from simple heuristics to advanced metaheuristics, machine learning techniques, RL, and DRL strategies \cite{machineLearningControl,heuristic_metaheurisitics}. In simpler production environments, basic heuristics such as FCFS, SPT, Earliest Due Date (EDD), and Cost Over Time (COVERT) work reasonably well \cite{fcfs_spt}. These heuristics can often be adapted for slightly more complex settings, though they may not be universally applicable. Metaheuristics, on the other hand, tend to offer more robust solutions and can handle greater system complexity. Approaches such as Genetic Algorithms (GA), Artificial Bee Colony (ABC), Simulated Annealing (SA), Tabu Search (TS), and Particle Swarm Optimization (PSO) have been widely used for various control tasks in production environments \cite{metaheurisitics_survey}. However, as system complexity increases, particularly when subjected to more complex dynamics and uncertainties, these methods may struggle.

Recent developments in machine learning, particularly RL and DRL, have become prominent for controlling complex systems with complex dynamics and uncertainties. These methods have been thoroughly explored in the literature. For instance, some studies have applied RL-based Q-learning to optimize robot assignments in multi-product Flexible Manufacturing Systems (FMS) \cite{waseem2024dynamic,Muhammad_ASME}, while others have used DRL-based actor-critic methods for optimizing robot scheduling and assignments \cite{bhatta2024dynamic}. Additionally, DRL has been used to manage machine status in systems with varying cycle times \cite{chen2022hybrid}. These RL/DRL approaches excel in handling system complexities and uncertainties, such as machine disruptions or fluctuating customer orders. However, a significant challenge with these models is their “black-box” nature, which makes them difficult to interpret and explain.

Although some studies have explored the issue of explainability in RL and DRL models, the term itself is multifaceted \cite{explainable_RL}. It can encompass aspects like transparency, informativeness, confidence, causality, interactivity, and transferability, among others. Different audiences may require different types of explanations, where experts might need a more technical explanation, non-experts, such as shop floor operators, require simpler, more understandable reasoning for decision-making. A comprehensive review of these explainability challenges in DRL is provided by \cite{explainable_RL}. Most explainability methods target expert users, while few address the needs of non-expert operators. Yet, in real-world manufacturing environments, non-expert users are often the ones who need to understand why specific decisions are taken or why certain events occur.

Another challenge with RL and DRL models is their scalability and resource demands. While these models can perform well in specific environments, scaling them to larger, more complex systems is often difficult. This scaling requires retraining the models, which not only demands expert knowledge to adapt the models but also requires significant computational resources \cite{li2023deep}.

In recent years, attention has shifted to LLMs, which are making significant strides in various fields, including manufacturing \cite{LLM2024embodied,llm2023enabling}. Models like OpenAI's GPT series, possess advanced pattern recognition capabilities, allowing them to process vast amounts of unstructured data and generate valuable insights across diverse tasks. Leading manufacturing companies, such as GE and Siemens, are actively exploring LLMs tailored to their specific needs. For example, Siemens has integrated LLMs into their Polarion software to enhance workflows, using AI for tasks such as data import, classification, and content generation \cite{siemen}. However, the application of LLMs in manufacturing remains in its early stages, with most existing studies focusing on simpler tasks. For instance, some studies have used ChatGPT to generate code for robotic pick-and-place tasks \cite{paiZheng_pickPlcae,paizheng2024vision}, while others have explored the use of LLMs as assistants in manufacturing \cite{paiZheng_llmAsst}. However, to our knowledge, no studies have explored the use of LLMs for real-time control to date.

Therefore this study proposes a novel control framework that utilizes pretrained LLMs for real-time control in serial production lines. The framework’s effectiveness is evaluated by comparing its performance to existing control policies such as FCFS, SPT, LPT, and DRL-based approaches

\subsection{Background}
\begin{enumerate}
    \item \textbf{LLM:} LLMs are advanced next-token prediction models trained on massive datasets, enabling them to perform a wide range of tasks with remarkable adaptability. LLMs are built upon the Transformer architecture, introduced by Vaswani et al. in 2017 \cite{vaswani2017attention}, which revolutionized natural language processing. At the core of the Transformer is the multi-head attention mechanism, which utilizes Query (Q), Key (K), Value (V), and Weight (W) vectors to determine the contextual significance of elements within an input sequence. This powerful mechanism allows LLMs to capture complex linguistic relationships, making them highly versatile and effective across various domains.

The Transformer’s decoder operates iteratively, predicting the next token in a sequence by analyzing the preceding tokens and selecting the most probable outcome at each step. To further refine their outputs, LLMs employ advanced techniques such as nucleus sampling and chain-of-thought reasoning, which enhance both the coherence and contextual accuracy of generated content. These capabilities position LLMs as valuable tools for diverse applications, including manufacturing, where they excel in processing unstructured data, supporting decision-making, and adapting to new tasks without extensive retraining on domain-specific knowledge.


\item \textbf{MARL Algorithms:} MARL, particularly Multi-Agent Actor-Critic algorithms, as referred to in this paper, is a family of reinforcement learning algorithms that consist of two key components: the actor, which determines the action to be taken, and the critic, which evaluates how effective the action is \cite{Zhang2021}. The actor learns the optimal policy by directly updating the policy parameters, while the critic assesses the quality of the policy by computing a value function. In this context, each agent learns a policy $\pi(u^i|\sigma^i)$, and the critic evaluates the policy based on the global state of the system.

The actor adjusts its policy parameters using a gradient that depends on the value function computed by the critic. The gradient for the Multi-Agent Actor-Critic algorithm can be expressed as:  
\begin{equation}
    \nabla_{\theta}J = \mathbf{E}_\pi\Bigg[ \sum_i \nabla_{\theta} \log \pi(u^i|\sigma^i) \big(Q(s,\mathbf{u}) - V(s)\big)\Bigg]
\end{equation}
where $V(s)$ is the value function computed by the critic, and $Q(s,\mathbf{u})$ is the state-action value estimated as:  
\begin{equation}
    Q(s,\mathbf{u}) = r + \gamma V(s),
\end{equation}  
with $\gamma$ representing the discount factor. Throughout this paper, the term MARL specifically refers to Multi-Agent Actor-Critic algorithms.

\end{enumerate}

\subsection{System Description}
\begin{figure}
    \centering
    \includegraphics[width=0.8\linewidth]{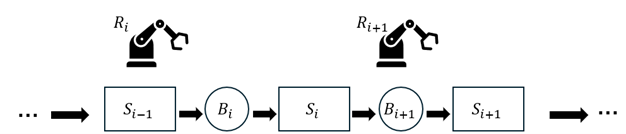}
    \caption{A general robot assisted serial production line}
    \label{fig:man_plant}
\end{figure}

The system described in this paper is a serial manufacturing line incorporating robots for material transport. These robots are responsible for transferring parts between multiple workstations. However, the challenge arises from having fewer robots than workstations, necessitating a strategic approach to efficiently load and unload parts at the various workstations. An illustration of this system is presented in Fig.~\ref{fig:man_plant}, where the robots are denoted by $R_i$ ($i = 1, 2, \dots, r$), workstations by $S_i$ ($i = 1, 2, \dots, M$), and buffers by $B_i$ ($i = 2, 3, \dots, M$). The cycle time for the $i^{\text{th}}$ workstation is represented as $T_i$, while the buffer capacity is denoted as $B_i$, with a slight abuse of notation. The model for the manufacturing system is based on our previous study \cite{waseem2024nash}. However, to keep the paper self contained, we will provide a general overview of the manufacturing system model here. The dynamics of the manufacturing system can be described by the following state-space equation:
\begin{equation}
    \Dot{\textbf{\textit{X}}} (t)=\textbf{\textit{F}}(\textbf{\textit{X}}(t),\textbf{\textit{U}}(t),\textbf{\textit{W}}(t))
\end{equation}
where 
\begin{itemize}
	\item $\textbf{\textit{X}}(t)=[X_1 (t),X_2 (t),\dots,X_M (t)]'$ represents the accumulated production count of each machine $S_i$ upto time $t$.
    
	\item $\textbf{\textit{U}}(t)=[u_1 (t),u_2(t),\dots,u_M (t)]'$ represents the control inputs at time $t$. Here, $u_i(t)\in\{0,1\}$ indicates whether machine $S_i$ is assigned a robot or not.
    
	\item $\textbf{\textit{W}}(t) = [w_1 (t), w_2 (t),\dots, w_M (t)]'$ represents the disturbances at time $t$, where $w_j(t)$ describes whether $S_j$  suffers from a disruption at time $t$. Specifically, if a machine is disrupted, $w_i (t)=1$, otherwise, $w_i(t)=0$. 
    
\end{itemize}

The accumulated production difference between two machines $S_i$ and $S_j$, $i,j\in 1,2,\dots,M,i\neq j$, within the time period $[0,t]$, follows the conservation of flow, which could be represented with the following equations:

\begin{equation}
X_i (t) - X_j (t) = \tau_{ij} (t) = \begin{cases}
\sum_{k=j+1}^{i}  b_k (0) - \sum_{k=j+1}^{i} b_k (t), \quad i > j\\
\\
\sum_{k=i+1}^{j}  b_k (t) - \sum_{k=i+1}^{j} b_k (0), \quad i < j
\end{cases}
\end{equation}

where $b_k(t)$ is the buffer level of the $k^{th}$ workstation at time $t$. The production difference $\tau_{ij} (t)$ are bounded by the condition that all buffers between machine $S_i$ and $S_j$ are full (for $i<j$) or empty (for $i>j$). Denote the boundary as $\beta_{ij} (t)$.

\begin{equation}
\beta_{ij} (t)= \begin{cases}
\sum_{k=j+1}^i ~b_k (0), ~~~~~~~~~~~~~~~~~~~~~i>j\\
\\
\sum_{k=i+1}^j ~B_k - \sum_{k=i+1}^j b_k (0), ~i<j
\end{cases}
\end{equation}

Thus, $\tau_{ij} (t) \leq \beta_{ij} (t)$.  Considering the interactions between $S_i$ and $S_j$, in the case of $\tau_{ij} (t)<\beta_{ij} (t)$, machine $S_i$ is not starved or blocked by $S_j$, thus it will process parts at its own rated speed. If $\tau_{ij} (t)=\beta_{ij} (t)$, the processing speed of machine $S_i$ will be constrained by machine $S_j$. Define a segment function $\xi(u,v)$ as

\begin{equation}
    \xi(u,v)= \begin{cases}
    +\infty,~~u<0 \\
    v, ~~~~~~~u=0
    \end{cases}
\end{equation}

Then, the actual process speed of machine $S_i$ can be described as:

\begin{equation}
    \dot{X}_l(t) = \min \left\lbrace \frac{\xi \left( (X_i(t) - X_j(t)) - \beta_{ij} , ~ u_j(t) \left( 1 - w_j(t) \right) \right)}{T_j(t)}, \frac{u_i(t) \left( 1 - w_i(t) \right)}{T_i(t)} \right\rbrace
    \label{eq:1}
\end{equation}

Extending this equation to all machines in the system, we have

\[
\dot{X}_i(t) = 
\min \left\{
\begin{array}{l}
\frac{\xi \left( (X_i(t) - X_1(t)) - \beta_{i1} \right), \, u_1(t) (1 - W_1(t))}{T_1(t)} \\
\frac{\xi \left( (X_i(t) - X_2(t)) - \beta_{i2} \right), \, u_2(t) (1 - W_2(t))}{T_2(t)} \\
~~~~~~~~~~~~~~~~~~~~~~~\vdots \\
~~~~~~~~~~~~~~\frac{u_i(t) (1 - W_i(t))}{T_i(t)} \\
~~~~~~~~~~~~~~~~~~~~~~~\vdots \\
\frac{\xi \left( (X_i(t) - X_M(t)) - \beta_{iM} \right), \, u_M(t) (1 - W_M(t))}{T_M(t)} \\
\end{array}
\right\}
\]

\begin{equation}
= f_i(X(t), U(t), W(t))
\end{equation}

Thus, the state-space function of the production count could be summarized as:

\begin{equation}
    \mathbf{X}(t) = \begin{bmatrix}
    F_1(\mathbf{X}(t), \mathbf{U}(t), \mathbf{W}(t)) \\
    \vdots \\
    F_M(\mathbf{X}(t), \mathbf{U}(t), \mathbf{W}(t))
    \end{bmatrix} = \mathbf{F}(\mathbf{X}(t), \mathbf{U}(t), \mathbf{W}(t))
    \label{eq:system_dynamics}
\end{equation}

\begin{equation}
\textbf{\textit{Y}}(t) = X_M(t) = [0 \ldots 0 \, 1] \textbf{\textit{X}}(t) = \textbf{\textit{H}}(\textbf{\textit{X}}(t))
\label{eq:placeholder}
\end{equation}

where $\textbf{Y}(t)$ is the production output of the last machine $S_M$ at time $t$, also known as system throughput.

The buffer levels of a certain buffer $b_{i+1}$ at time $t$ could be calculated as: 

\begin{equation}
b_{i+1} (t)=X_i (t)-X_{i+1} (t)+b_{i+1}(0)
\end{equation}

If sensor data for random disruption events, machine inputs $u_1 (0),u_2 (0),…,u_M (0)$ and initial buffer levels $b_1 (0),b_2 (0),… ,b_{M-1} (0)$ are provided, the detailed state of the system at any given time can be recursively evaluated by (4)-(11).

\section{Problem Formulation and Proposed Framework}

\begin{figure}[t]
    \centering
    \includegraphics[width=1\linewidth]{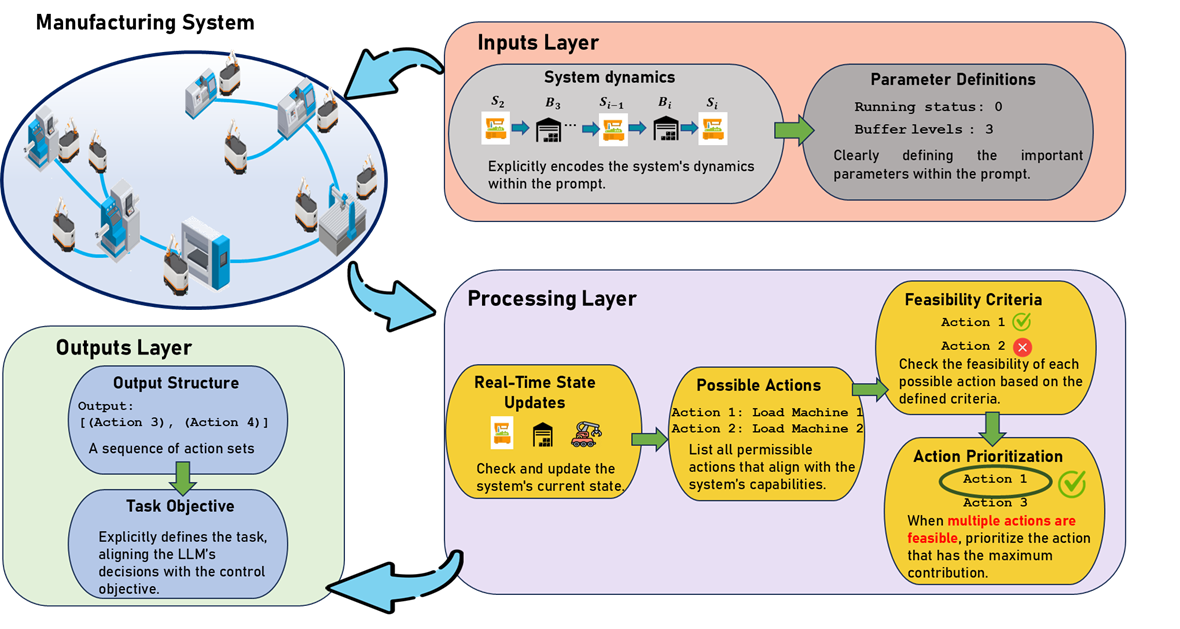}
    \caption{Proposed LLM-based control framework}
    \label{fig:fig2}
\end{figure}

Efficient control of manufacturing systems is crucial for achieving high throughput and operational efficiency. The complexity of this task stems from several factors: First, the dynamic behavior of the system plays a key role. Machines and buffers operate under constantly changing conditions influenced by factors like processing times, disruptions, and resource allocation. For example, sudden machine failures or buffer overflows can disrupt the production flow, requiring quick and effective control decisions. Second, real-time decision-making is essential, especially when assigning mobile robots for material handling. These decisions need to be made almost instantaneously to prevent bottlenecks and optimize throughput, taking into account both the current state of the system and the feasibility of potential actions.

In this section we discuss our proposed framework which leverages pre-trained LLMs to directly control manufacturing operations using structured natural language prompts. This approach eliminates the need for retraining and reduces computational costs while providing interpretable, scalable, and adaptable decision-making. The proposed framework is structured into three distinct layers: the input layer, the processing layer, and the output layer. A summary of these layers is provided below.

\subsection{Input Layer}
    
    This layer provides essential information about the system that the LLM will control. All the data in this layer is static for the system under consideration and includes system dynamics, along with various parameters and terminologies relevant to the system.
\begin{enumerate}
    \item  \textit{System Dynamics}: This detailed description enables the LLM to grasp the underlying structure of the production line and its interactions. It specifies the system dynamics, including the number of machines, buffers, material handling systems, and their interconnections. By providing this information, the LLM can accurately determine the appropriate actions to take. Additionally, this description is easily modifiable by non-experts, allowing for straightforward adjustments if the system needs to be altered.

\item\textit{Parameter Definitions}: This description defines the terminology used in the prompt. Since different production systems involve varying parameters that contain unique and critical information about the system’s operation, explicitly stating all relevant parameters is crucial for generating effective actions in the system under study.

\end{enumerate}

\subsection{Processing Layer}

This layer provides the necessary information for the LLM to make decisions and process data effectively. It includes real-time state information, potential actions, feasibility criteria, and the prioritization of actions.

\begin{enumerate}

    \item \textit{Real-Time State Updates}: This part of the prompt conveys information about the current state of the system. It is crucial that the state information provided is thorough enough to enable the LLM to make informed decisions about the appropriate actions. For instance, in a manufacturing context, relevant details such as the operational status of machines, robot positions, and buffer statuses must be included.

    \item \textit{Possible Actions}: This part of the prompt defines the feasible actions in the current state of the system. In real-time control, it’s vital to take actions that are both appropriate and recognized by the system. To prevent the issue of "hallucination"—where the LLM produces illogical outputs—it’s crucial to clearly specify and define each action, ensuring the generated outputs are both meaningful and actionable.

    \item \textit{Feasibility Criteria}: This part of the prompt defines the feasibility criteria for each possible action. Since LLMs are prone to hallucinations and may select actions arbitrarily without fully considering the system's dynamics, it is crucial to explicitly inform the LLM about the feasibility of each action based on the current system state. 

    \item \textit{Action Prioritization}: This part of the prompt addresses the prioritization of feasible actions. In complex, dynamic systems, multiple actions may be feasible simultaneously, making the role of the control algorithm crucial in selecting the most appropriate action. To optimize system performance, actions should be chosen based on their potential to achieve the desired control objective, such as maximizing throughput. Unlike RL/DRL approaches, where policies are trained to identify optimal actions, LLMs are not specifically trained for this task. Therefore, to guide the LLM toward optimal decisions, it is essential to clearly define the priorities of available actions. Action prioritization must be explicitly incorporated into the prompt, providing the LLM with instructions on how to evaluate and rank actions based on their relevance and impact on system performance.
\end{enumerate}

\subsection{Output Layer}

The output layer enables the LLM to understand how to present the requested task, including aspects such as format, structure, and size.

\begin{enumerate}

    \item \textit{Output Structure}: This part of the prompt specifies the output format. Since the output generated by the LLM is directly integrated into the manufacturing system, it is crucial that the generations adhere to a precise and predefined format. Even small deviations can lead to significant errors or disruptions. To ensure the LLM produces valid and effective actions, the output structure must be clearly defined, with examples provided to illustrate the correct format. Additionally, the LLM should be informed of the expected number of actions it needs to generate and the steps to take if no action is feasible.


    \item \textit{Task Objective}: This final part of the prompt defines the LLM's task and provides guardrails to prevent hallucinations. It clearly outlines the intended outcome, ensuring that the LLM understands its task and remains focused on generating accurate, relevant information. Guardrail statements are included to prevent the generation of incorrect or irrelevant responses.
    

\end{enumerate}

This well-structured framework proves to be an innovative approach to real-time manufacturing system control. By explicitly encoding system dynamics and constraints into the prompt, the framework operates without the need for environment-specific training or retraining, making it highly adaptable to evolving production line configurations. This structured approach allows immediate reconfiguration for system changes, such as the addition of new machines or buffers, enabling seamless scalability across diverse manufacturing environments.

A core advantage of our framework is its focus on ensuring logical, actionable decisions through robust feasibility criteria. By explicitly defining the conditions under which actions are valid, the framework maintains decision reliability in real-time operations. This structured decision-making process mitigates risks often associated with LLM outputs, such as generating infeasible or ambiguous actions, ensuring consistent and effective control even in complex, dynamic scenarios.

The framework also emphasizes interpretability, producing human-readable outputs that detail the decision-making process in an understandable format. This transparency enhances usability, allowing operators to trust and verify the system's actions without requiring specialized expertise. Coupled with minimal computational overhead, the proposed methodology demonstrates a practical and scalable solution for real-time manufacturing control, with a clear focus on operational efficiency and ease of deployment.

\section{Numerical Study}
In this section, numerical experiments are conducted to assess the effectiveness of LLM-based control for real-time mobile robot scheduling. A serial production line with two configurations is studied: (1) a simple configuration of two machines, one buffer, and a single mobile robot; and (2) a more complex configuration of four machines, three buffers, and two mobile robots. For comparison, several alternative approaches such as MARL, FCFS, SPT, and LPT are applied to solve the mobile robot assignment problem. System throughput is used as the primary performance metric. The comparison is based on a single episode of 480 time steps (equivalent to 8 hours of operation per day), replicated 25 times to obtain the standard deviation. From these case studies, two key conclusions are drawn: (1) the proposed LLM-based control scheme effectively optimizes mobile robot scheduling; and (2) it outperforms the FCFS, SPT, and LPT heuristics in terms of system throughput, while performing comparably to RL/MARL-based control. Additionally, it offers the advantage of providing clear, explainable rationale for each decision (action) made. 

This section is organized into five subsections. The first subsection outlines the parameters for the two tested configurations. The second subsection explains the performance metric used and its significance. The third subsection introduces the benchmark policies against which our approach is evaluated. The fourth subsection presents the experimental results, while the fifth subsection provides a detailed discussion of these results.

\subsection{Configuration Parameters}

The experiments utilize synthetic data derived from real-world automobile industry data. While confidentiality restrictions prevent the use of actual data, the synthetic data closely mirrors real scenarios, making it suitable for simulation, analysis, training, and demonstration purposes.

\begin{enumerate}
    \item \textbf{Configuration 1}: This configurations consists of two machines, one buffer and one mobile robot. Both the machines have the same processing time of $3$ minutes and a constant load and unload time of $2$ minutes, buffer capacity of $B_i = 3$ units, and the initial buffer level of zero. Downtime is not considered for this configuration.

    \item \textbf{Configuration 2}: This configuration comprises four machines, three intermediate buffers, and two mobile robots. Information about the machines used in configuration 2 is provided in Table 1. The loading and unloading time for each machine is constant, set at $2$ minutes. All buffers have a fixed capacity of $B_i=3$ units, with initial buffer levels assumed to be zero. The machines in this configuration are subject to random downtime, which follows an exponential distribution. The Mean Time Between Failures (MTBF) and Mean Time To Repair (MTTR) for the machines are provided in Table 1.

\end{enumerate}

\begin{table}[h]
\centering
\caption{Machines parameters for configuration 2.}
\begin{tabular}{|c|c|c|c|}
\hline
\textbf{Machines, $S_i$} & \textbf{Processing time (min)} & \textbf{MTBF (min)} & \textbf{MTTR (min)} \\ \hline
$S_1$ & 3  & 100 & 20  \\ \hline
$S_2$ & 4  & 120 & 30  \\ \hline
$S_3$ & 3  & 125 & 25  \\ \hline
$S_4$ & 3  & 150 & 25  \\ \hline
\end{tabular}
\end{table}

For both configurations, the LLM-based controller utilized prompts corresponding to each part of the input, processing, and output layers. Below, we present the prompts used for Configuration 2. It is important to note that the prompts for Configuration 1 follow the same structure.

\begin{itemize}
    \item \textbf{System dynamics:} \textit{Consider a serial manufacturing system with four machines (Machines 1–4) and three in-between buffers, each with a maximum capacity of 3 parts. Each machine has dedicated processing time, with Machine 2 being the slowest (it takes longer to process a part than the others). Two robots are in the system to handle materials.}
    \item \textbf{Parameter definitions:} \textit{"running status" indicates if a machine is up or down: 0 = down, 1 = up;
"mp-status" indicates if a machine is loaded with a part: 0 = machine has no part, 1 = machine has a part;
"progress" indicates if a machine is processing/has processed a part: 0 = either part processing is completed or no part on the machine at all, 0 < value < 1 = part being processed;
"robot status" indicates if a robot is assigned to a machine: 0 = not assigned, 1 = assigned.}

    \item \textbf{Real-time state:} \textit{Machine 1 running status: \{state[0]\};
Machine 2 running status: \{state[1]\};
Machine 3 running status: \{state[2]\};
Machine 4 running status: \{state[3]\};
Machine 1 mp-status: \{state[4]\};
Machine 2 mp-status: \{state[5]\};
Machine 3 mp-status: \{state[6]\};
Machine 4 mp-status: \{state[7]\};
Machine 1 progress: \{state[8]\};
Machine 2 progress: \{state[9]\};
Machine 3 progress: \{state[10]\};
Machine 4 progress: \{state[11]\};
Buffer 1 level: \{state[12]\};
Buffer 2 level: \{state[13]\};
Buffer 3 level: \{state[14]\};
Machine 1 robot status: \{state[15]\};
Machine 2 robot status: \{state[16]\};
Machine 3 robot status: \{state[17]\};
Machine 4 robot status: \{state[18]\}.}

    \item \textbf{Possible actions:}\textit{Action 1 (0,0): Load machine 1;  
Action 2 (0,1): Load machine 2;  
Action 3 (0,2): Load machine 3;  
Action 4 (0,3): Load machine 4.}

    \item \textbf{Feasibility criteria:}\textit{ Action 1 (Load Machine 1):  
     Feasible if:  
        Buffer 1 level < 3;  
        Machine 1 robot status is 0 (not assigned);  
        Machine 1 is running (running status = 1); 
        Either:  
            Machine 1 mp-status is 0 (not loaded), or  
            Machine 1 mp-status is 1 and progress is 0 (part has been processed and ready to be unloaded.)}

    \item \textbf{Action prioritization:} \textit{If multiple actions are feasible, they must be prioritized as follows:
  Action 4 (0,3) has the highest priority if feasible;
  Action 3 (0,2) comes next in priority;
  Action 2 (0,1) should be prioritized less than Action 3;
  Action 1 (0,0) is the least prioritized action.}

    \item \textbf{Output structure:} \textit{If only one action is feasible, it must be accompanied by Action 1 (0,0). For example, if only Action 3 is feasible, the output should be: `[(0,2), (0,0)]`.
If multiple actions are feasible, output the actions in prioritized order.
If no actions are feasible, return `[(0,0), (0,0)]` to indicate that no machine can be loaded.}

    \item \textbf{Task objective:} \textit{Based on these states, criteria, and the goal of maximizing production throughput, what actions should be assigned to the two robots? The output must be in the format `[(action1), (action2)]` without additional details. The feasibility criteria must be strictly followed.}
\end{itemize}

\subsection{Performance Metrics}
The primary performance metric considered in this study is system throughput, which is defined as the total production count of the last machine in the serial production line within a given operational period. Throughput is a direct measure of the system's efficiency and serves as an indicator of the effectiveness of the control strategy applied. It reflects the combined performance of all machines, buffers, and robots in the production line. Any inefficiency, such as bottlenecks, idle time, or disruptions, directly reduces throughput, making it a comprehensive measure of system control effectiveness. In practical applications, high throughput ensures that production goals are met, resource utilization is optimized, and operational costs are minimized.

\subsection{Benchmark Policies}

To evaluate the proposed LLM-based control framework, its performance is compared with the following benchmark policies, each representing different approaches to robot scheduling and system control:

\begin{enumerate}
    \item \textbf{FCFS}: It is a straightforward scheduling policy in which tasks are processed in the exact order they arrive, without any regard for priority or processing time. Under this approach, each part entering the system is fully processed before the next part is introduced. In the context of a manufacturing system, this means that robots are assigned to machines sequentially, starting from the first machine and proceeding in order until the last machine is reached, ensuring that each task is completed before the next one begins. However, it is inefficient for systems with varying machine speeds or dynamic disruptions, as it fails to prioritize critical tasks or address bottlenecks.

    \item \textbf{SPT}: It is a scheduling policy that prioritizes tasks according to their processing time, with the task requiring the shortest processing time being processed first. In this approach, robots are assigned to machines based on the shortest processing times. If multiple machines have the same processing time, robots are assigned in sequential order, moving along the production line to ensure each machine is utilized efficiently. However, it may neglect slower machines, potentially creating bottlenecks in downstream processes.
    
\item \textbf{LPT}: It is a scheduling policy that prioritizes tasks based on their processing times, with the task requiring the longest processing time being executed first. In this approach, robots are assigned to the machines with the longest processing times, meaning the slowest machines are given priority. If multiple machines have the same longest processing time, robots are assigned sequentially, following the order of machines along the production line. However, it may underutilize faster machines, leading to inefficiencies in certain scenarios.

\item \textbf{MARL}: MARL is an advanced control approach where robots are modeled as agents that learn to optimize scheduling decisions based on a predefined reward function, such as maximizing throughput. In manufacturing systems, MARL enables agents to interact with the environment and learn optimal policies. Each robot makes independent decisions about its next action based on the global system state, allowing the framework to handle complex dynamics and adapt to disruptions or varying operational conditions. However, MARL requires substantial computational resources for training, along with significant time and expertise for hyperparameter tuning and retraining. Furthermore, the "black-box" nature of MARL models poses challenges to explainability.

In our comparison, we specifically employ the Multi-Agent Proximal Policy Optimization (MAPPO) algorithm. Within our system, mobile robots responsible for material handling are treated as agents. The observations, actions, and rewards for each agent are defined as follows:

\begin{enumerate}
    \item \textbf{State/Observation}: The system is fully observable, meaning the state and observation are identical. Each agent observes the following variables at time \( t \):

    \begin{equation}
    o_i(t) = \{\theta_i(t), mp_i(t), mpg_i(t), b_i(t), r_i(t)\}
    \end{equation}

    Here:
    \begin{itemize}
        \item \( \theta_i(t) \): Operational status of machine \( i \) at time \( t \),
        \item \( mp_i(t) \): Part status of machine \( i \) (\( mp_i(t) = 1 \) if the machine has a part; \( mp_i(t) = 0 \) otherwise),
        \item \( mpg_i(t) \): Machining progress of the part at machine \( i \) at time \( t \),
        \item \( b_i(t) \): Buffer level of buffer \( i \) at time \( t \),
        \item \( r_i(t) \): Robot assignment status for machine \( i \) (\( r_i(t) = 1 \) if a robot is assigned to the machine; \( r_i(t) = 0 \) otherwise).
    \end{itemize}

    \item \textbf{Action}: The action for each robot (agent) is defined as the index of the machine to which the robot is assigned:

    \begin{equation}
    a_i(t) = S_i
    \end{equation}

    where \( S_i \) represents the index of the machine assigned to the robot.

    \item \textbf{Reward}: The reward for each agent is defined based on the system throughput:

    \begin{equation}
    R_i(t) = Y(t)
    \end{equation}

    where \( Y(t) \) represents the system throughput, measured as the production count of the last machine.
\end{enumerate}

\end{enumerate}

\subsection{Results}

   \begin{figure}[t!]
    \centering
    \includegraphics[width=0.8\linewidth]{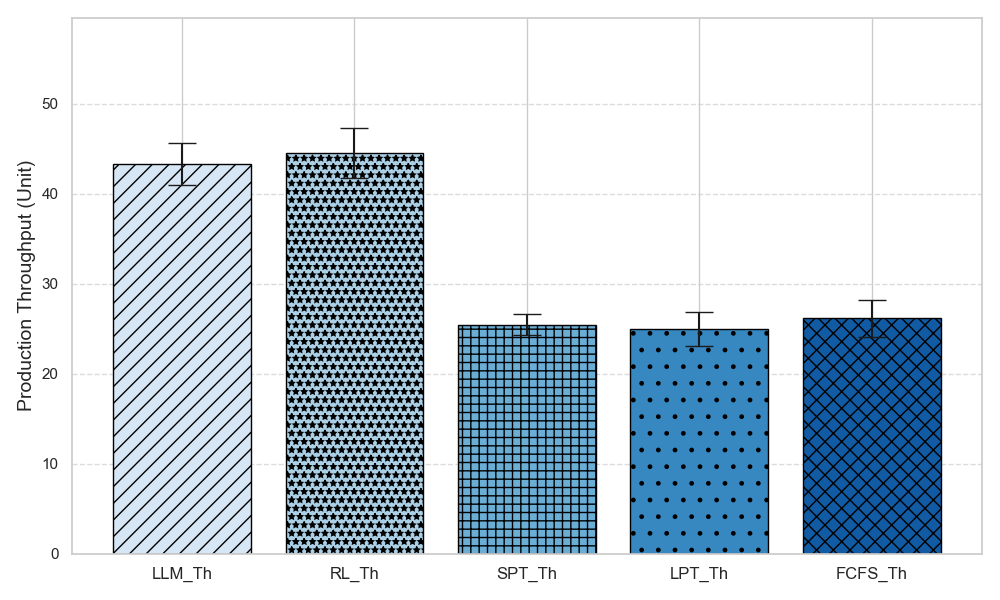}
    \caption{Performance comparison based on configuration 1}
    \label{fig:config_1_results}
\end{figure}

 The performance of our proposed framework on configuration 1 is shown in Fig.~\ref{fig:config_1_results}. As shown, our approach closely mirrors the performance of the RL-based control, with a mean throughput difference of only 1-2 units. Both control policies achieve an average throughput of approximately $45$ units. Although RL-based control is typically expected to outperform other methods due to its explicit training on the environment, the LLM-based approach performs nearly identically in this configuration. This can be attributed to the simplicity of the environment, which consists of only two machines and a single robot. Given the relatively straightforward decision-making process, the LLM model achieves near-optimal performance with minimal complexity. On the other hand, while the rule-based controls such as SPT, LPT, and FCFS demonstrate lower performance compared to the RL and LLM-based control policies, they exhibit identical performance with a mean throughput of approximately $25$ units. This uniformity in performance is primarily due to the synchronous nature of the system, where both machines have similar cycle times. Consequently, both SPT and LPT tend to prioritize Machine 1, followed by Machine 2, which aligns closely with the sequence followed by FCFS.

\begin{figure}
    \centering
    \includegraphics[width=0.8\linewidth]{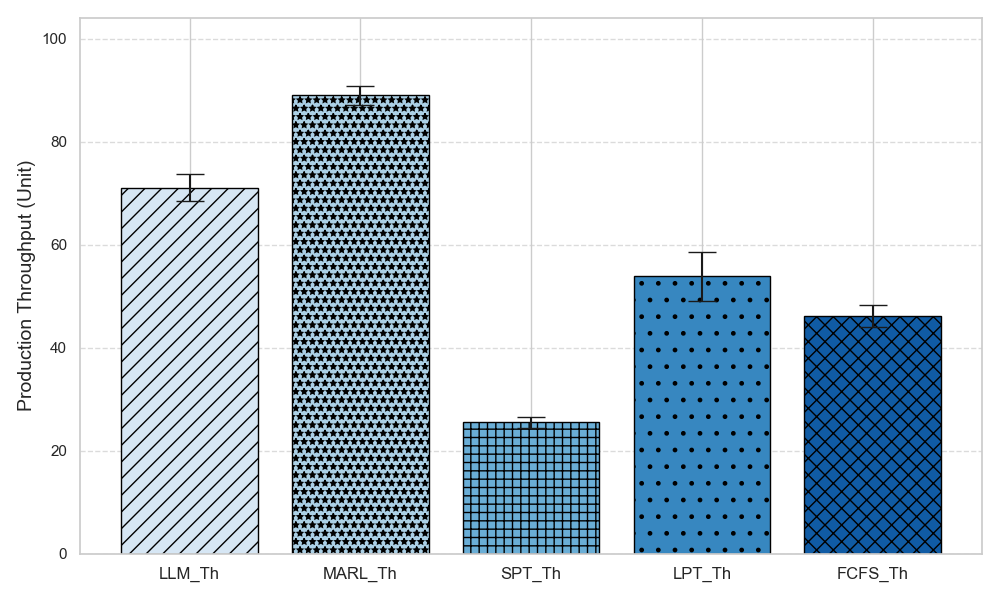}
    \caption{Performance comparison based on configuration 2}
    \label{fig:config_2_results}
\end{figure}

Fig.~\ref{fig:config_2_results} presents a comparison of production throughput based on Configuration 2. As shown, the MARL-based control yields the highest average throughput of 89 units, with a very low standard deviation. While the proposed LLM-based control does not achieve the same level of performance as the MARL-based control, this outcome was anticipated. The MARL-based control benefits from extensive training and continuous interaction with the environment, which justifies its superior performance. In contrast, the proposed LLM-based control relies solely on prompt-based natural language instructions, requiring no training and being computationally efficient. When compared to other baselines, the proposed LLM-based control demonstrates a strong performance, highlighting its potential as an effective and practical alternative.

\begin{figure}[t]
    \centering
    \includegraphics[width=1\linewidth]{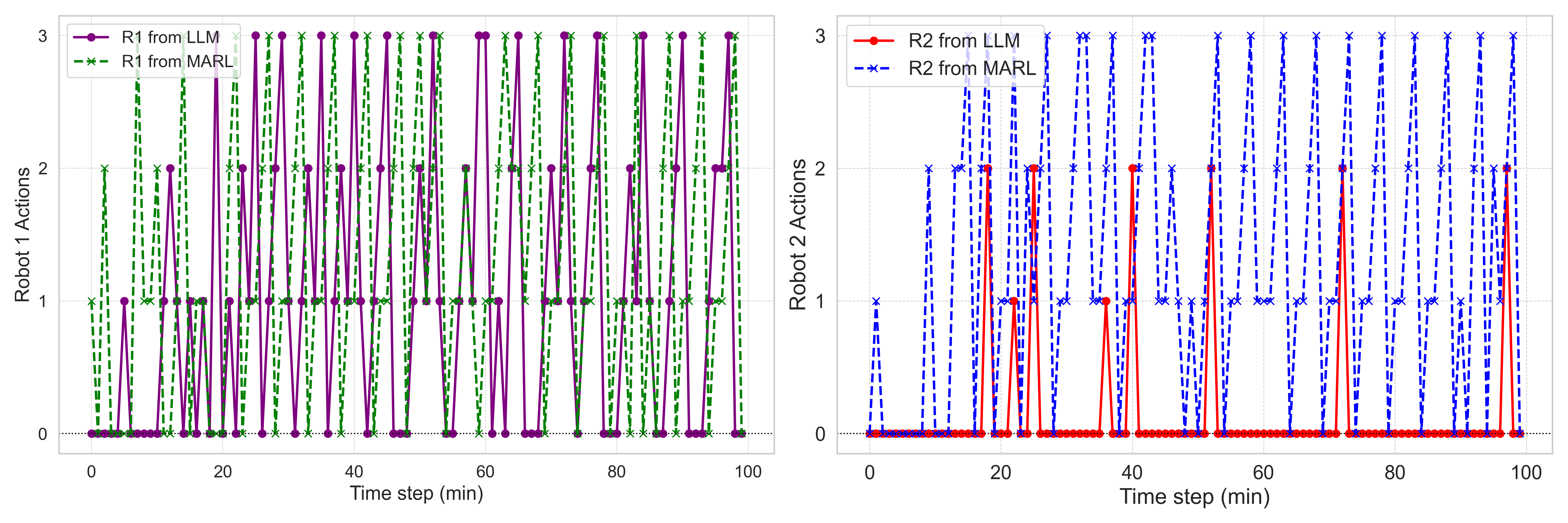}
    \caption{Actions comparison from LLM and MARL}
    \label{fig:actions_comp}
\end{figure}

To further investigate the performance difference between LLM-based control and MARL-based control, we compare the actions selected and the buffer levels variation under the trained MARL policy and the instructed LLM. Fig.~\ref{fig:actions_comp} compares the actions taken by the LLM and MARL over the first $100$ timesteps. This subset of the timeline is considered for clarity, as including the full 480-minute period would create congestion, making it harder to read. The actions of Robot 1 and Robot 2 are shown separately. The y-axis represents the machine number assigned to each robot (i.e., 0, 1, 2, or 3, corresponding to the four machines).

In the first $20$ minutes, both robots are primarily assigned to Machine 0, as reflected by the predominantly zero values in the action data. This is due to two factors: firstly, the system starts with zero buffer levels, and Machine 0 is the only machine capable of receiving an unlimited number of parts. Consequently, the LLM assigns both robots to Machine 0, adhering to the feasibility criteria. Secondly, when no feasible actions are available, both robots are kept at Machine 0, as it is the only machine that can accommodate any robot at the start. In the case of MARL, while most actions are similarly assigned to Machine 0 during this period, some actions are assigned to other machines. However, these alternate assignments are rejected by the environment due to ineligibility, which does not alter the overall behavior.

For the remainder of the timeline, the actions of Robot 1 from both LLM and MARL align closely, following a similar structure with only minor deviations, mainly due to the initial actions. For Robot 2, LLM predominantly assigns it to Machine 0 or leaves it unused. It is rarely used for loading/unloading tasks for other machines, and during this 100-step period, it is never assigned to Machine 3. Instead, Robot 1 is more frequently assigned to Machine 3.

In contrast, MARL demonstrates a more balanced distribution of actions between both robots. This balance is a result of MARL’s thorough exploration of the environment, which allows it to understand the consequences of its actions in different states. By comparison, the LLM follows a more rigid, rule-based approach, relying heavily on a single robot for most tasks and only utilizing the second robot when absolutely necessary. Furthermore, due to its strict adherence to pre-defined rules, the LLM does not take actions that might benefit it in the long term. Nevertheless, despite these differences, the overall performance of LLM closely mirrors that of MARL.

\begin{figure}
    \centering
    \includegraphics[width=1\linewidth]{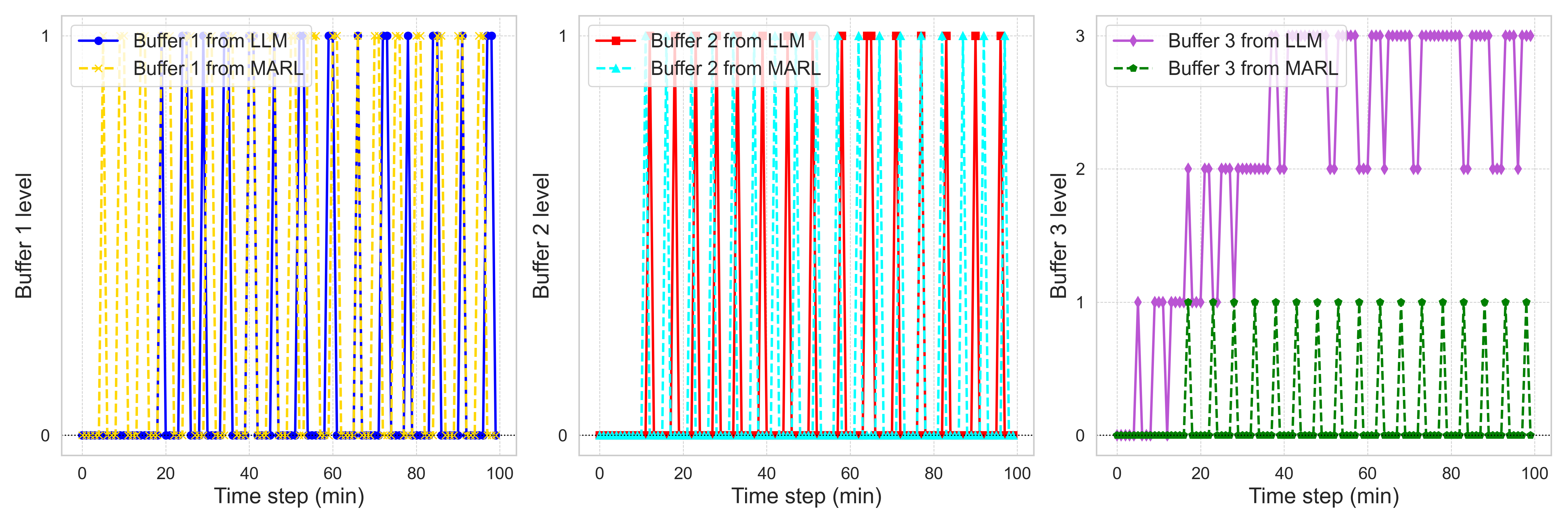}
    \caption{Buffer level variation under MARL and LLM based control}
    \label{fig:buffer_comp}
\end{figure}

Fig.~\ref{fig:buffer_comp} illustrates the variations in buffer levels for the three intermediate buffers. The data is shown for the first 100 minutes to avoid graph congestion and ensure clarity. Each buffer has a maximum capacity of 3 units and the y-axis represents the number of units in each buffer at each timestep, ranging from 0 to 3.

Starting with Buffer 1, it is evident that under the LLM control policy, the buffer level remains empty, especially during the first 20 minutes. In contrast, the MARL-based control policy consistently maintains Buffer 1 at level 1. After the initial 20 minutes, both the MARL and LLM policies align more closely, with the buffer level staying at 1 for the remainder of the observed period.

For Buffer 2, both LLM and MARL follow nearly identical trends. This similarity may be attributed to the influence of the preceding machine, Machine 1, which feeds into Buffer 2. While MARL learns this behavior through its interaction with the environment, the LLM is explicitly instructed to avoid conditions such as blockage and starvation. As a result, both policies aim to maintain a similar level of operation, which ensures that Buffer 2 is kept at a steady state.

In the case of Buffer 3, the trends under the two control policies diverge. MARL maintains Buffer 3 at level 1, similar to the other buffers, while the LLM policy attempts to increase the buffer level. This behavior stems from the fact that Buffer 3 is directly linked to the final machine in the system, which determines the overall system throughput. To prevent the last machine from being starved, the LLM policy prioritizes filling Buffer 3 to ensure a smooth flow of parts and mitigate any potential risks to throughput. This explains why the LLM starts filling Buffer 3 during the first 20 minutes, which in turn contributes to the emptiness observed in Buffer 1 during this period. The LLM assigns the parts to Buffer 3 and holds them there to ensure that the downstream machine has enough work to continue operating. This approach also aligns with the action selection behavior under both the LLM and MARL policies, where each policy seeks to achieve a similar overall goal but through different strategies.

\subsection{Discussion}
Based on the above study, it is evident that RL-based control outperforms the proposed LLM-based control as well as the other baselines. This raises the question: why opt for LLM-based control when MARL-based control performs better? The following sections outline the key reasons for choosing LLM-based control over MARL-based control.

\begin{enumerate}
\item \textbf{Explainability}

Machine learning and deep learning have demonstrated great utility in various applications, yet they often remain "black-box" models, leaving users with little insight into the reasoning behind a decision. While there has been ongoing research aimed at improving the interpretability of these models, they still pose significant challenges, particularly for non-expert users who may struggle to understand the rationale behind automated decisions. In contrast, LLMs offer a compelling advantage in this regard by providing transparent and understandable explanations for their actions. Not only can LLMs generate clear justifications for their decisions, but they also enable users to engage in reasoning about the decisions themselves. This ability to reason allows for deeper insights and verification of the model's behavior.

\begin{figure}
    \centering
    \includegraphics[width=1\linewidth]{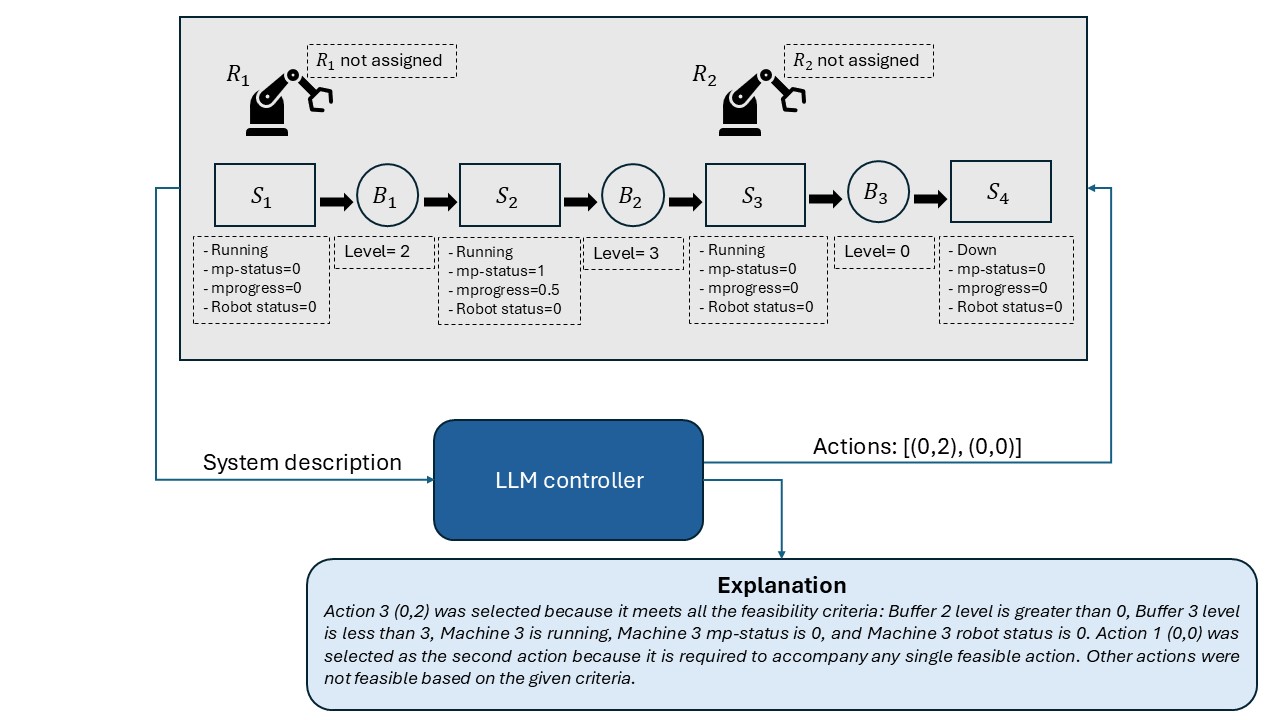}
    \caption{Action output along with explanation from the proposed LLM-based control}
    \label{fig:explanation}
\end{figure}

LLMs provide intuitive and human-readable explanations, detailing the factors influencing their decision-making process. For instance, as illustrated in Figure 7, when the current state of the system is fed into the LLM, the model not only generates an action but also explains why that particular action was selected, in comparison to other potential options. This transparency empowers users to understand the underlying logic of the model and provides confidence in the system's decisions, a capability that remains elusive in many other machine learning paradigms. 
 \\
\item \textbf{Computational efficiency}

A significant limitation of RL and DRL approaches is their inherent computational complexity. These models require extensive training to converge to an optimal policy, a process that involves interacting with the environment and adjusting the model’s parameters over numerous episodes. While this may not be an issue in relatively simple environments with small state and action spaces, it becomes a formidable challenge as the complexity of the system grows. For instance, in our current scenario, where the environment includes only four machines, three intermediate buffers, two mobile robots for material handling, and a single product type, the training process still takes approximately two days to complete. This training time is manageable for small-scale systems, but when scaling up to more complex systems, featuring additional machines, buffers, robots, and multiple product types, the computational cost grows exponentially, leading to increased time and resource consumption.

In contrast, LLM-based control offers a significant advantage in terms of computational efficiency. Unlike RL/DRL models, LLMs do not require extensive training or continuous interaction with the environment. Instead, they leverage their pre-trained knowledge, which can be easily adapted to new tasks through the provision of a well-crafted prompt. As demonstrated in our proposed framework, the process involves describing system parameters such as machines, buffers, robots, and product types to the LLM, which then generates decisions based on this input, without the need for retraining. This approach substantially reduces the computational burden and eliminates the need for time-consuming training cycles.

Moreover, this method empowers shop floor operators, who may not be domain experts, to update the system and generate decisions simply by adjusting the system’s parameters in the prompt. As a result, LLM-based control is not only more computationally efficient but also more accessible and scalable compared to traditional RL/DRL-based methods. RL/DRL models typically require a large number of trial-and-error interactions to collect sufficient data and learn from it, which can be both computationally expensive and time-consuming, particularly in complex environments. This issue is exacerbated in systems where obtaining new samples or interactions is costly, impractical, or time-consuming. In manufacturing systems, for instance, each interaction may correspond to an entire day of operations or involve complex physical resources, making the required number of samples for model training prohibitive.

Furthermore, RL/DRL models often necessitate the careful adjustment of numerous hyperparameters to ensure optimal performance. This tuning process can be both time-consuming and computationally intensive, often relying on trial and error, grid search, or other optimization techniques. In large and complex environments, the number of parameters to adjust grows, further complicating the task of achieving optimal performance.
\\

\item \textbf{Generalization}

One of the key challenges in traditional RL and DRL methods is their limited ability to generalize across different environments or problem scenarios. RL models, in particular, tend to overfit to the specific environment they are trained on. This means that once a model has been trained for a particular system, it may not perform well when applied to a different system with even slight variations, such as a change in the number of machines, robots, or products, or a modification in the system’s layout or workflow. This lack of generalization is particularly problematic in dynamic or evolving environments, where frequent retraining would be required to adapt the model to new conditions, leading to significant overhead in terms of both time and computational resources.

In contrast, LLM-based control offers superior generalization capabilities. Since LLMs are pre-trained on large-scale data and possess a vast amount of knowledge, they are inherently more adaptable to a wide range of tasks and environments without requiring extensive retraining. With proper prompts, an LLM can handle new scenarios or tasks by simply adjusting the system's parameters in the prompt, allowing the model to generalize to different operational contexts. For instance, changes in the number of machines, robots, or product types can be accommodated by modifying the input description provided to the LLM, rather than retraining the model from scratch. This flexibility enables LLM-based control to seamlessly adapt to a variety of system configurations and operational changes, making it highly scalable and versatile.

Additionally, LLMs can leverage their ability to reason and contextualize information to handle scenarios that have not been explicitly encountered before, provided that the prompt sufficiently represents the system's dynamics. This ability to reason and adapt without needing exhaustive retraining makes LLMs an ideal choice for environments where variability and complexity are frequent, and where the time and resource costs of retraining traditional RL models would be prohibitive.
As the complexity of the system increases (e.g., more machines, buffers, or robots), LLMs simply require updating the input prompt with new parameters, without the need for retraining the model. This eliminates the need for continuous model fine-tuning, making the system easier to scale.
Moreover, LLM-based control is low-maintenance because it operates based on a pre-trained model. Once configured, it doesn’t require constant retraining or performance evaluations, unlike RL/DRL models, which need frequent adjustments to maintain efficiency as the system evolves.
\\

\item \textbf{User-friendly}

RL/DRL models typically require an in-depth understanding of the algorithm’s architecture, hyperparameters, and the system dynamics in order to effectively set up, train, and fine-tune the model. Moreover, the process of configuring and modifying these models to accommodate different environments or tasks is often time-consuming and error-prone, requiring specialized knowledge in both machine learning and the domain of application. This barrier to entry limits the adoption of RL/DRL in many real-world settings, especially in operational environments where domain experts may not possess the technical skills required to work with complex machine learning models.
In contrast, LLM-based control is inherently more user-friendly, primarily due to its simplicity and accessibility. Rather than requiring complex model training or hyperparameter tuning, the user is only required to provide a well-structured prompt describing the system's parameters and desired objectives. The system operates based on this input, without the need for technical expertise in machine learning. The ability to easily adjust the system's behavior by updating the prompt further enhances its usability, allowing non-experts, such as shop floor operators or business managers, to directly interact with and control the system. This level of accessibility ensures that the technology can be easily deployed across different industries without requiring a deep understanding of machine learning principles.
Additionally, LLMs offer a high degree of interpretability, as users can interact with the system in natural language and receive explanations for its decisions. This human-readable interface not only facilitates decision-making but also allows users to trust the system's actions, knowing they can ask for clarification on the reasoning behind any decision. This makes LLM-based control not only more accessible but also more transparent, improving the overall user experience and making it easier to integrate into existing workflows.
\\

\item \textbf{Integration with external knowledge}

Unlike RL and DRL models, which rely heavily on the data generated within the environment for learning, LLMs can incorporate external sources of information, such as domain expertise, industry guidelines, or historical data. This allows the system to make informed decisions based not only on the current state of the environment but also on a broader context.
For example, LLMs can be prompted to consider best practices or specific rules that may not be part of the training data but are relevant for the decision-making process. This capability enhances the flexibility and accuracy of LLM-based control, enabling it to adapt to complex, real-world scenarios where additional knowledge is crucial for optimal performance.
\\
\item \textbf{Cost-effective and customizable}

RL/DRL typically requires substantial computational resources for training, especially in complex environments. This includes the need for large-scale simulations, extensive training data, and frequent retraining to maintain performance. These costs can quickly accumulate, especially in dynamic or large-scale systems.
On the other hand, LLM-based control eliminates the need for time-consuming training and high computational costs. Once an LLM is pre-trained, it can be directly applied to the control task with minimal computational overhead. This makes it an attractive option for environments where computational resources are limited, or cost-effectiveness is a priority. Additionally, the simplicity of LLMs reduces the need for continuous model updates, further cutting down long-term costs.
With RL/DRL, customization typically requires reconfiguring or retraining the model, which can be time-consuming and computationally expensive, especially as system requirements evolve.

\end{enumerate}

\section{Conclusion}
In this paper, we propose an LLM-based control framework for real-time management of a serial production line. Unlike traditional control methods that rely on fixed rules or training, our approach utilizes the pretrained GPT-4 model to make dynamic control decisions. The framework consists of three layers: an input layer, a processing layer, and an output layer. The input layer provides static system information, such as configurations, dynamics, and entity relationships. The processing layer handles dynamic information, including the current system status, possible actions, feasibility criteria, and action prioritization, which may change over time. The output layer generates decisions based on the input and processing layers and outputs them in a predefined format. This structure ensures that the LLM model understands both the requirements and the rationale behind its decisions. The framework is tested using production throughput as the key performance metric. The results show that it outperforms traditional control methods such as FCFS, SPT, and LPT. However, as anticipated, it does not achieve the same level of performance as MADRL-based control, since MADRL models are fine-tuned for specific environments, while the pre-trained LLM has not been customized for this task. Nevertheless, the proposed framework stands out for its explainability, offering greater transparency into the decision-making process.

As generative models, LLMs can sometimes produce unexpected output. In this study, despite a structured prompt, the LLM occasionally deviates from the expected format. Although such errors are minor in many cases, they can be problematic in real-time systems, where misformatted outputs may lead to system failures. Addressing this issue is crucial for ensuring smooth operation in real-world applications.

\section*{Acknowledgements}
This work is supported by National Science Foundation Grants 2243930.

\bibliographystyle{unsrt}
\bibliography{llm_reference}

\end{document}